\documentclass[runningheads]{llncs}
\usepackage[T1]{fontenc}
\usepackage{graphicx}
\usepackage{amsmath,amssymb} 
\usepackage{amsmath,epsfig}
\usepackage{graphicx}
\usepackage{bm}
\usepackage{amssymb}
\usepackage{cite}
\usepackage{array}
\usepackage{extarrows}
\usepackage{url}
\usepackage{float}
\usepackage{multirow}
\usepackage{booktabs}
\usepackage{blkarray}
\usepackage{multirow}
\usepackage{array}
\usepackage{color}
\begin{document}
\title{Fully Transformer Network for Change Detection of Remote Sensing Images}
%
%
\author{Tianyu Yan \and
Zifu Wan \and
Pingping Zhang\orcidID{0000-0003-1206-1444}}
\authorrunning{T. Yan et al.}
%
\institute{School of Artificial Intelligence, Dalian University of Technology, China
\email{\{tianyuyan2001,wanzifu2000\}@gmail.com;zhpp@dlut.edu.cn}}
\maketitle              
\begin{abstract}
Recently, change detection (CD) of remote sensing images have achieved great progress with the advances of deep learning.
However, current methods generally deliver incomplete CD regions and irregular CD boundaries due to the limited representation ability of the extracted visual features.
To relieve these issues, in this work we propose a novel learning framework named Fully Transformer Network (FTN) for remote sensing image CD, which improves the feature extraction from a global view and combines multi-level visual features in a pyramid manner.
More specifically, the proposed framework first utilizes the advantages of Transformers in long-range dependency modeling.
It can help to learn more discriminative global-level features and obtain complete CD regions.
Then, we introduce a pyramid structure to aggregate multi-level visual features from Transformers for feature enhancement.
The pyramid structure grafted with a Progressive Attention Module (PAM) can improve the feature representation ability with additional interdependencies through channel attentions.
Finally, to better train the framework, we utilize the deeply-supervised learning with multiple boundary-aware loss functions.
Extensive experiments demonstrate that our proposed method achieves a new state-of-the-art performance on four public CD benchmarks.
For model reproduction, the source code is released at https://github.com/AI-Zhpp/FTN.

\keywords{Fully Transformer Network  \and Change Detection \and Remote Sensing Image.}
\end{abstract}
\section{Introduction}
Change Detection (CD) plays an important role in the field of remote sensing.
It aims to detect the key change regions in dual-phase remote sensing images captured at different times but over the same area.
Remote sensing image CD has been used in many real-world applications, such as land-use planning, urban expansion management, geological disaster monitoring, ecological environment protection.
However, due to change regions can be any shapes in complex scenarios, there are still many challenges for high-accuracy CD.
In addition, remote sensing image CD by handcrafted methods is time-consuming and labor-intensive, thus there is a great need for fully-automatic and highly-efficient CD.

In recent years, deep learning has been widely used in remote sensing image processing due to its powerful feature representation capabilities, and has shown great potential in CD.
With deep Convolutional Neural Networks (CNN)~\cite{ioffe2015batch,he2016deep,huang2017densely}, many CD methods extract discriminative features and have demonstrated good CD performances.
However, previous methods still have the following shortcomings: 1) With the resolution improvement of remote sensing images, rich semantic information contained in high-resolution images is not fully utilized. As a result, current CD methods are unable to distinguish pseudo changes such as shadow, vegetation and sunshine in sensitive areas. 2) Boundary information in complex remote sensing images is often missing. In previous methods, the extracted changed areas often have regional holes and their boundaries can be very irregular, resulting in a poor visual effect~\cite{liu2020building}. 3) The temporal information contained in dual-phase remote sensing images is not fully utilized, which is also one of the reasons for the low performance of current CD methods.

To tackle above issues, in this work we propose a novel learning framework named Fully Transformer Network (FTN) for remote sensing image CD, which improves the feature extraction from a global view and combines multi-level visual features in a pyramid manner.
More specifically, the proposed framework is a three-branch structure whose input is a dual-phase remote sensing image pair.
We first utilize the advantages of Transformers~\cite{vaswani2017attention,dosovitskiy2020image,liu2021swin} in long-range dependency modeling to learn more discriminative global-level features.
Then, to highlight the change regions, the summation features and difference features are generated by directly comparing the temporal features of dual-phase remote sensing images.
Thus, one can obtain complete CD regions.
To improve the boundary perception ability, we further introduce a pyramid structure to aggregate multi-level visual features from Transformers.
The pyramid structure grafted with a Progressive Attention Module (PAM) can improve the feature representation ability with additional interdependencies through channel attentions.
Finally, to better train the framework, we utilize the deeply-supervised learning with multiple boundary-aware loss functions.
Extensive experiments show that our method achieves a new state-of-the-art performance on four public CD benchmarks.

In summary, the main contributions of this work are as follow:
\begin{itemize}
\item We propose a novel learning framework (\emph{i.e.}, FTN) for remote sensing image CD, which can improve the feature extraction from a global view and combine multi-level visual features in a pyramid manner.
\item We propose a pyramid structure grafted with a Progressive Attention Module (PAM) to further improve the feature representation ability with additional interdependencies through channel attentions.
\item We introduce the deeply-supervised learning with multiple boundary-aware loss functions, to address the irregular boundary problem in CD.
\item Extensive experiments on four public CD benchmarks demonstrate that our framework attains better performances than most state-of-the-art methods.
\end{itemize}
\section{Related Work}
\subsection{Change Detection of Remote Sensing Images}
Technically, the task of change detection takes dual-phase remote sensing images as inputs, and predicts the change regions of the same area.
Before deep learning, direct classification based methods witness the great progress in CD.
For example, Change Vector Analysis (CVA)~\cite{huo2009fast,xiaolu2011change} is powerful in extracting pixel-level features and is widely utilized in CD.
With the rapid improvement in image resolution, more details of objects have been recorded in remote sensing images.
Therefore, many object-aware methods are proposed to improve the CD performance.
For example, Tang \emph{et al.}~\cite{tang2011object} propose an object-oriented CD method based on the Kolmogorov--Smirnov test.
Li \emph{et al.}~\cite{li2016change} propose the object-oriented CVA to reduce the number of pseudo detection pixels.
With multiple classifiers and multi-scale uncertainty analysis, Tan \emph{et al.}~\cite{tan2019object} build an object-based approach for complex scene CD.
Although above methods can generate CD maps from dual-phase remote sensing images, they generally deliver incomplete CD regions and irregular CD boundaries due to the limited representation ability of the extracted visual features.

With the advances of deep learning, many works improve the CD performance by extracting more discriminative features.
For example, Zhang~\emph{et al.}~\cite{zhang2016feature} utilize a Deep Belief Network (DBN) to extract deep features and represent the change regions by patch differences.
Saha~\emph{et al.}~\cite{saha2019unsupervised} combine a pre-trained deep CNN and traditional CVA to generate certain change regions.
Hou~\emph{et al.}~\cite{hou2017change} take the advantages of deep features and introduce the low rank analysis to improve the CD results.
Peng~\emph{et al.}~\cite{peng2019unsupervised} utilize saliency detection analysis and pre-trained deep networks to achieve unsupervised CD.
Since change regions may appear any places, Lei~\emph{et al.}~\cite{lei2019multiscale} integrate Stacked Denoising AutoEncoders (SDAE) with the multi-scale superpixel segmentation to realize superpixel-based CD.
Similarly, Lv~\emph{et al.}~\cite{lv2018deep} utilize a Stacked Contractive AutoEncoder (SCAE) to extract temporal change features from superpixels, then adopt a clustering method to produce CD maps.
Meanwhile, some methods formulate the CD task into a binary image segmentation task.
Thus, CD can be finished in a supervised manner.
For example, Alcantarilla~\emph{et al.}~\cite{alcantarilla2018street} first concatenate dual-phase images as one image with six channels.
Then, the six-channel image is fed into a Fully Convolutional Network (FCN) to realize the CD.
Similarly, Peng~\emph{et al.}~\cite{peng2019end} combine bi-temporal remote sensing images as one input, which is then fed into a modified U-Net++~\cite{zhou2018unet++} for CD.
Daudt~\emph{et al.}~\cite{daudt2018fully} utilize Siamese networks to extract features for each remote sensing image, then predict the CD maps with fused features.
The experimental results prove the efficiency of Siamese networks.
Further more, Guo~\emph{et al.}~\cite{guo2018learning} use a fully convolutional Siamese network with a contrastive loss to measure the change regions.
Zhang~\emph{et al.}~\cite{zhang2020deeply} propose a deeply-supervised image fusion network for CD.
There are also some works focused on specific object CD.
For example, Liu~\emph{et al.}~\cite{liu2020building} propose a dual-task constrained deep Siamese convolutional network for building CD.
Jiang~\emph{et al.}~\cite{jiang2020pga} propose a pyramid feature-based attention-guided Siamese network for building CD.
Lei~\emph{et al.}~\cite{lei2020hierarchical} propose a hierarchical paired channel fusion network for street scene CD.
The aforementioned methods have shown great success in feature learning for CD.
However, these methods have limited global representation capabilities and usually focus on local regions of changed objects.
We find that Transformers have strong characteristics in extracting global features.
Thus, different from previous works, we take the advantages of Transformers, and propose a new learning framework for more discriminative feature representations.
\subsection{Vision Transformers for Change Detection}
Recently, Transformers~\cite{vaswani2017attention} have been applied to many computer vision tasks, such as image classification~\cite{dosovitskiy2020image,liu2021swin}, object detection~\cite{carion2020end}, semantic segmentation~\cite{wang2021end}, person re-identification~\cite{zhang2021hat,liu2021video} and so on.
Inspired by that, Zhang~\emph{et al.}~\cite{zhang2022swinsunet} deploy a Swin Transformer~\cite{liu2021swin} with a U-Net~\cite{ronneberger2015u} structure for remote sensing image CD.
Zheng~\emph{et al.}~\cite{zheng2022changemask} design a deep Multi-task Encoder-Transformer-Decoder (METD) architecture for semantic CD.
Wang~\emph{et al.}~\cite{wang2021transcd} incorporate a Siamese Vision Transformer (SViT) into a feature difference framework for CD.
To take the advantages of both Transformers and CNNs, Wang~\emph{et al.}~\cite{wang2022network} propose to combine a Transformer and a CNN for remote sensing image CD.
Li~\emph{et al.}~\cite{li2022transunetcd} propose an encoding-decoding hybrid framework for CD, which has the advantages of both Transformers and U-Net.
Bandara~\emph{et al.}~\cite{bandara2022transformer} unify hierarchically structured Transformer encoders with Multi-Layer Perception (MLP) decoders in a
Siamese network to efficiently render multi-scale long-range details for accurate CD.
Chen~\emph{et al.}~\cite{chen2021remote} propose a Bitemporal Image Transformer (BIT) to efficiently and effectively model contexts within the spatial-temporal domain for CD.
Ke~\emph{et al.}~\cite{ke2022hybrid} propose a hybrid Transformer with token aggregation for remote sensing image CD.
Song~\emph{et al.}~\cite{song2022mstdsnet} combine the multi-scale Swin Transformer and a deeply-supervised network for CD.
All these methods have shown that Transformers can model the inter-patch relations for strong feature representations.
However, these methods do not take the full abilities of Transformers in multi-level feature learning.
Different from existing Transformer-based CD methods, our proposed approach improves the feature extraction from a global view and combines multi-level visual features in a pyramid manner.
\begin{figure*}
\centering
\resizebox{1\textwidth}{!}
{
\begin{tabular}{@{}c@{}c@{}}
\includegraphics[width=1\linewidth,height=0.42\linewidth]{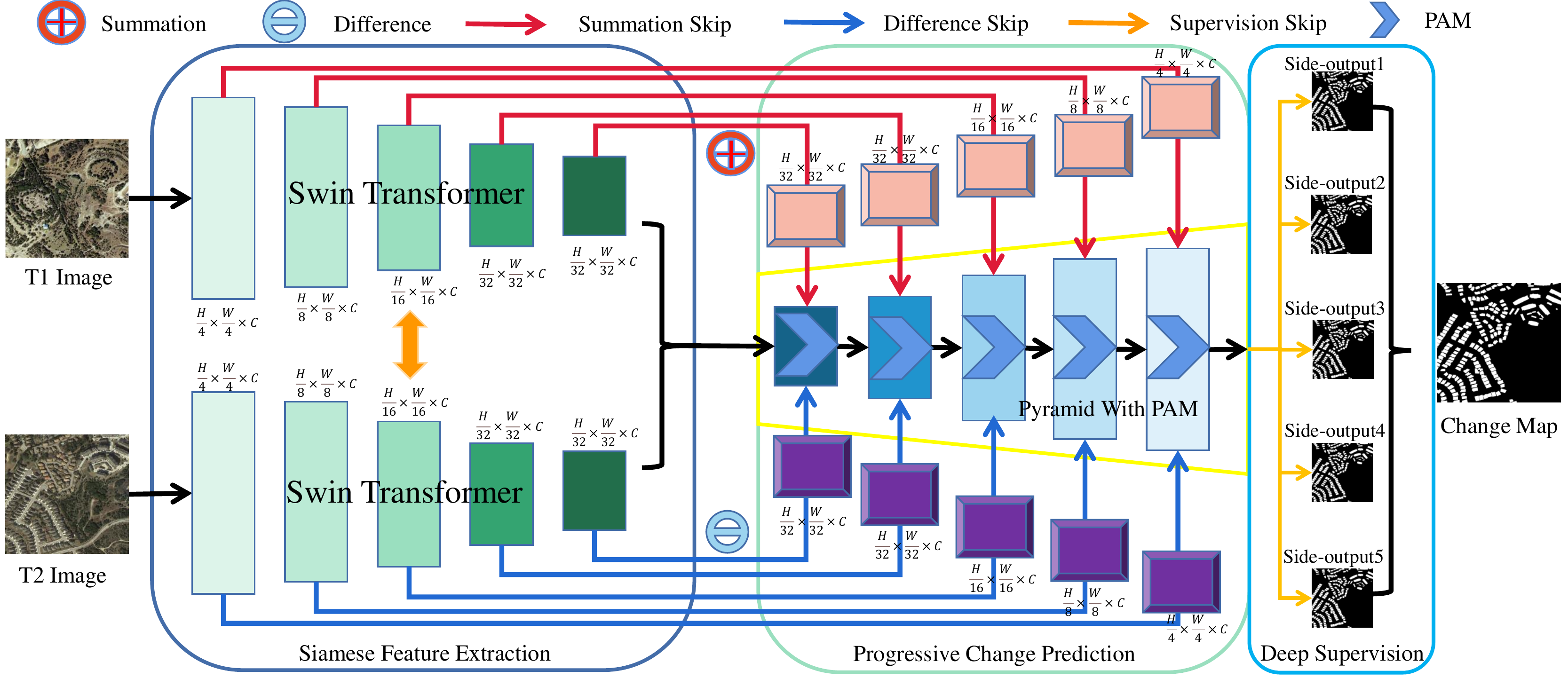} \\
\end{tabular}
}
\caption{The overall structure of our proposed framework.
}
\label{fig:Framework}
\end{figure*}
\section{Proposed Approach}
As shown in Fig.~\ref{fig:Framework}, the proposed framework includes three key components, \emph{i.e.}, Siamese Feature Extraction (SFE), Deep Feature Enhancement (DFE) and Progressive Change Prediction (PCP).
By taking dual-phase remote sensing images as inputs, SFE first extracts multi-level visual features through two shared Swin Transformers.
Then, DFE utilizes the multi-level visual features to generate summation features and difference features, which highlight the change regions with temporal information.
Finally, by integrating all above features, PCP introduces a pyramid structure grafted with a Progressive Attention Module (PAM) for the final CD prediction.
To train our framework, we introduce the deeply-supervised learning with multiple boundary-aware loss functions for each feature level.
We will elaborate these key modules in the following subsections.
\subsection{Siamese Feature Extraction}
Following previous works, we introduce a Siamese structure to extract multi-level features from the dual-phase remote sensing images.
More specifically, the Siamese structure contains two encoder branches, which share learnable weights and are used for the multi-level feature extraction of images at temporal phase 1 (T1) and temporal phase 2 (T2), respectively.
As shown in the left part of Fig.~\ref{fig:Framework}, we take the Swin Transformer~\cite{liu2021swin} as the basic backbone of the Siamese structure, which involves five stages in total.
\begin{figure*}
\centering
\resizebox{0.48\textwidth}{!}
{
\begin{tabular}{@{}c@{}c@{}}
\includegraphics[width=1\linewidth,height=0.82\linewidth]{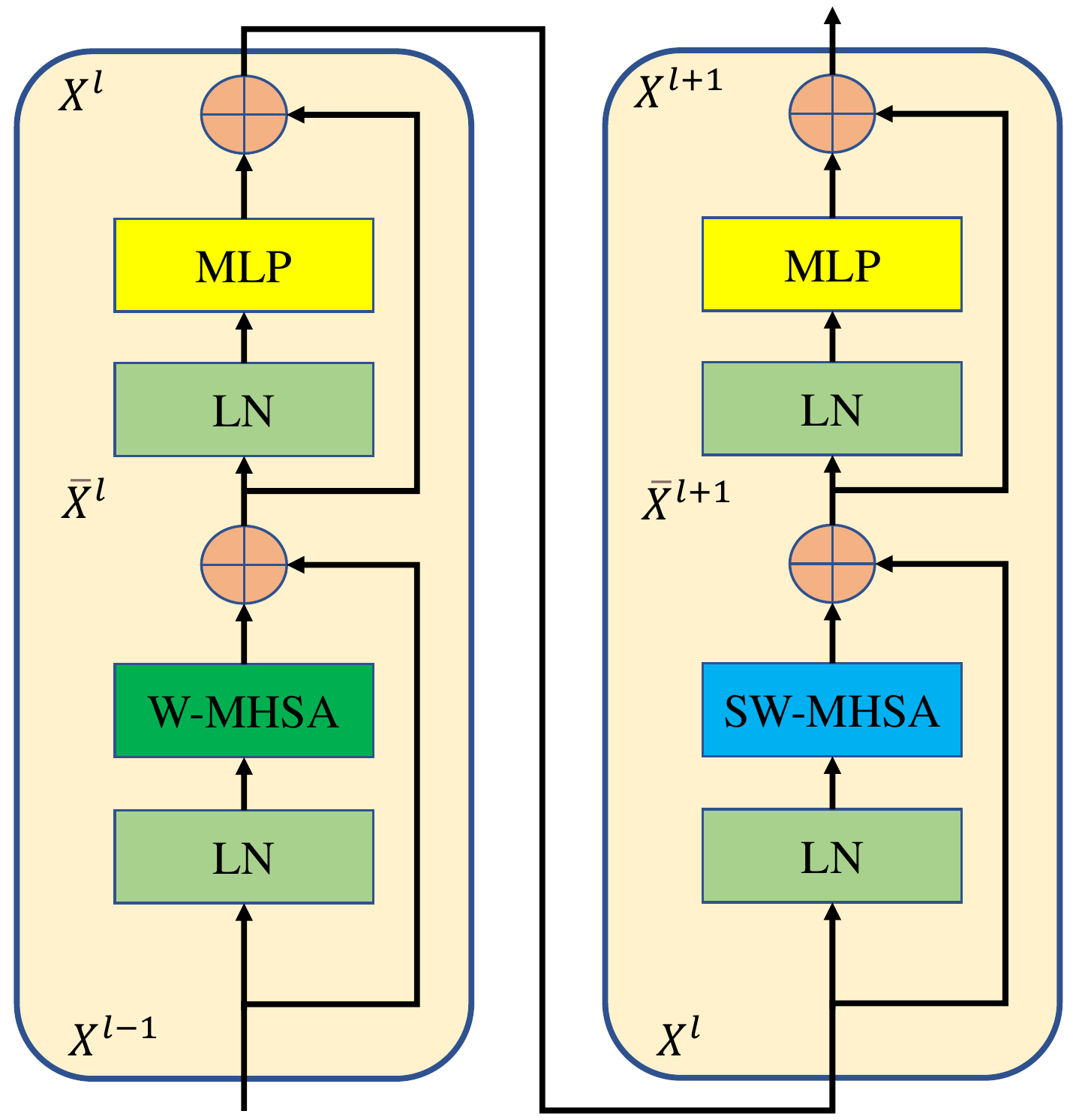} \\
\end{tabular}
}
\caption{The basic structure of the used Swin Transformer block.}
\label{fig:SwinT}
\end{figure*}

Different from other typical Transformers~\cite{vaswani2017attention,dosovitskiy2020image}, the Swin Transformer replaces the standard Multi-Head Self-Attention (MHSA) with Window-based Multi-Head Self-Attention (W-MHSA) and Shifted Window-based Multi-Head Self-Attention (SW-MHSA), to reduce the computational complexity of the global self-attention.
To improve the representation ability, the Swin Transformer also introduces MLP, LayerNorm (LN) and residual connections.
Fig.~\ref{fig:SwinT} shows the basic structure of the Swin Transformer block used in this work.
Technically, the calculation formulas of all the procedures are given as follows:
\begin{equation}\label{1}
\bar{\textbf{X}}^l = \text{W-MHSA}(\text{LN}(\textbf{X}^{l-1}))+\textbf{X}^{l-1},
\end{equation}
\begin{equation}\label{2}
\textbf{X}^l = \text{MLP}(\text{LN}(\bar{\textbf{X}}^{l-1}))+\bar{\textbf{X}}^{l},
\end{equation}
\begin{equation}\label{3}
\bar{\textbf{X}}^{l+1} = \text{SW-MHSA}(\text{LN}(\bar{\textbf{X}}^{l}))+\textbf{X}^{l},
\end{equation}
\begin{equation}\label{4}
\textbf{X}^{l+1} = \text{MLP}(\text{LN}(\bar{\textbf{X}}^{l+1}))+\bar{\textbf{X}}^{l+1},
\end{equation}
where $\bar{\textbf{X}}$ is the output of the W-MHSA or SW-MHSA module and $\textbf{X}$ is the output of the MLP module.
At each stage of the original Swin Transformers, the feature resolution is halved, while the channel dimension is doubled.
More specifically, the feature resolution is reduced from $(H/4)\times(W/4)$ to $(H/32)\times(W/32)$, and the channel dimension is increased from $C$ to $8C$.
In order to take advantages of global-level information, we introduce an additional Swin Transformer block to enlarge the receptive field of the feature maps.
Besides, to reduce the computation, we uniformly reduce the channel dimension to $C$, and generate encoded features $[\textbf{E}^{1}_{T1}, \textbf{E}^{2}_{T1},...,\textbf{E}^{5}_{T1}]$ and $[\textbf{E}^{1}_{T2}, \textbf{E}^{2}_{T2},...,\textbf{E}^{5}_{T2}]$ for the T1 and T2 images, respectively.
Based on the shared Swin Transformers, the multi-level visual features can be extracted.
In general, features in the high-level capture more global semantic information, while features in the low-level retain more local detail information.
Both of them help the detection of change regions.
\subsection{Deep Feature Enhancement}
In complex scenarios, there are many visual challenges for remote sensing image CD.
Thus, only depending on the above features is not enough.
To highlight the change regions, we propose to enhance the multi-level visual features with feature summation and difference, as shown in the top part and bottom part of Fig.~\ref{fig:Framework}.
More specifically, we first perform feature summation and difference, then introduce a contrast feature associated to
each local feature~\cite{luo2017non}.
The enhanced features can be represented as:
\begin{equation}\label{5}
\bar{\textbf{E}}^{k}_S = \text{ReLU}(\text{BN}(\text{Conv}(\textbf{E}^{k}_{T1}+\textbf{E}^{k}_{T2}))),
\end{equation}
\begin{equation}\label{6}
\textbf{E}^{k}_S = [\bar{\textbf{E}}^{k}_S, \bar{\textbf{E}}^{k}_S-\text{Pool}(\bar{\textbf{E}}^{k}_S)],
\end{equation}
\begin{equation}\label{7}
\bar{\textbf{E}}^{k}_D = \text{ReLU}(\text{BN}(\text{Conv}(\textbf{E}^{k}_{T1}-\textbf{E}^{k}_{T2}))),
\end{equation}
\begin{equation}\label{8}
\textbf{E}^{k}_D = [\bar{\textbf{E}}^{k}_D, \bar{\textbf{E}}^{k}_D-\text{Pool}(\bar{\textbf{E}}^{k}_D)],
\end{equation}
where $\textbf{E}^{k}_S$ and $\textbf{E}^{k}_D$ $(k=1,2,...,5)$ are the enhanced features with point-wise summation and difference, respectively.
$\text{ReLU}$ is the rectified linear unit, $\text{BN}$ is the batch normalization, $\text{Conv}$ is a $1\times1$ convolution, and $\text{Pool}$ is a $3\times3$ average pooling with padding=1 and stride=1.
[,] is the concatenation operation in channel.
Through the proposed DFE, change regions and boundaries are highlighted with temporal information.
Thus, the framework can make the extracted features more discriminative and obtain better CD results.
\subsection{Progressive Change Prediction}
\begin{figure*}
\centering
\resizebox{0.62\textwidth}{!}
{
\begin{tabular}{@{}c@{}c@{}}
\includegraphics[width=1\linewidth,height=0.44\linewidth]{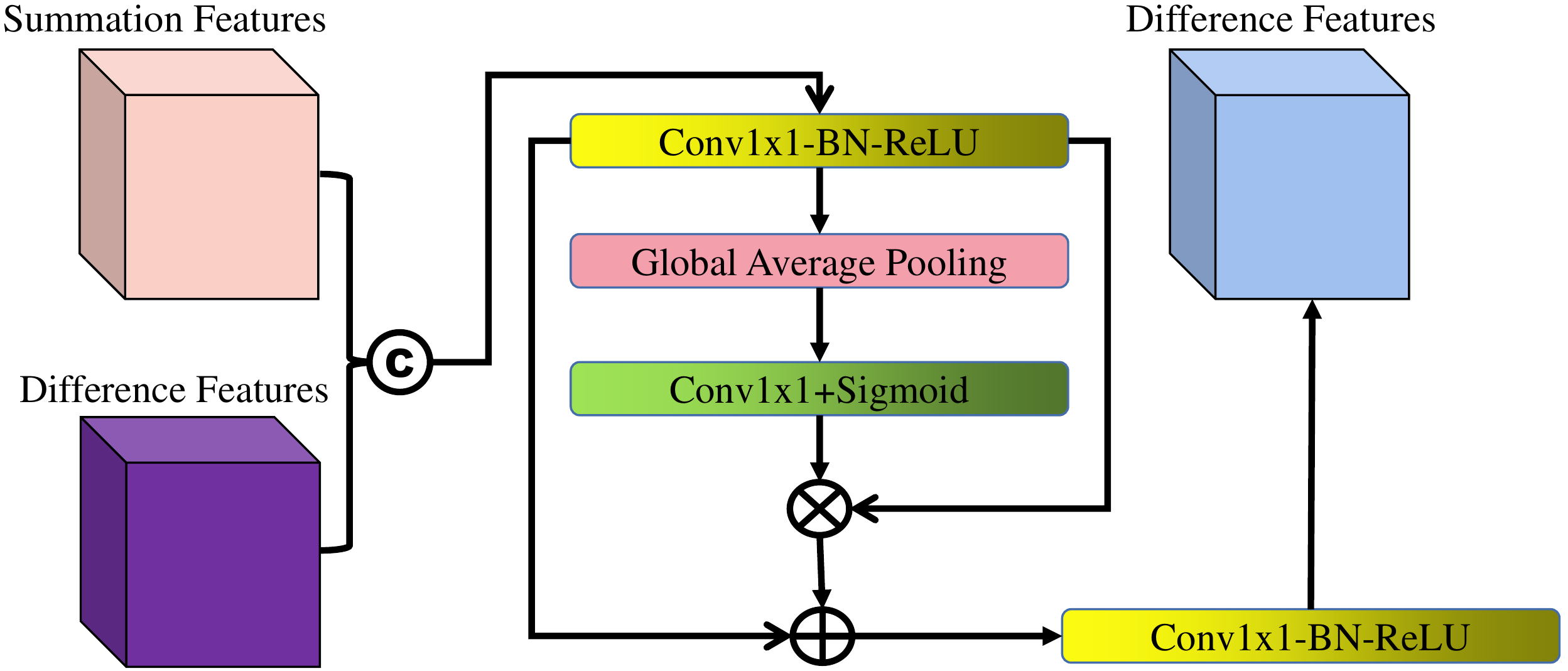} \\
\end{tabular}
}
\caption{The structure of our proposed Progressive Attention Module (PAM).}
\label{fig:PAM}
\end{figure*}

Since change regions can be any shapes and appear in any scales, we should consider the CD predictions at various cases.
Inspired by the feature pyramid~\cite{lin2017feature}, we propose a progressive change prediction, as shown in the middle part of Fig.~\ref{fig:Framework}.
To improve the representation ability, a pyramid structure with a Progressive Attention Module (PAM) is utilized with additional interdependencies through channel attentions.
The structure of the proposed PAM is illustrated in Fig.~\ref{fig:PAM}.
It first takes the summation features and difference features as inputs, then a channel-level attention is applied to enhance the features related to change regions.
Besides, we also introduce a residual connection to improve the learning ability.
The final feature map can be obtained by a $1\times1$ convolution.
Formally, the PAM can be represented as:
\begin{equation}\label{9}
\textbf{F}^{k} = \text{ReLU}(\text{BN}(\text{Conv}([\textbf{E}^{k}_S, \textbf{E}^{k}_D]))),
\end{equation}
\begin{equation}\label{10}
\textbf{F}^{k}_A = \textbf{F}^{k}*\sigma(\text{Conv}(\text{GAP}(\textbf{F}^{k})))+\textbf{F}^{k},
\end{equation}
where $\sigma$ is the Sigmoid function and $\text{GAP}$ is the global average pooling.

To achieve the progressive change prediction, we build the decoder pyramid grafted with a PAM as follows:
\begin{equation}\label{11}
\textbf{F}^{k}_P = \left\{
\begin{aligned}
&\textbf{F}^{k}_A,& \quad k=5,\\
&\text{UM}(\text{SwinBlock}^{n}(\textbf{F}^{k+1}_P)))+\textbf{F}^{k}_A,& \quad 1\le k<5.
\end{aligned}
\right.
\end{equation}
where $\text{UM}$ is the patch unmerging block used in Swin Transformers for upsampling, and $\text{SwinBlock}^{n}$ is the Swin Transformer block with $n$ layers.
From the above formula, one can see that our PCP can make full use of the interdependencies within channels, and can progressively aggregate multi-level visual features to improve the perception ability of the change regions.
\subsection{Loss Function}
To optimize our framework, we adopt the deeply-supervised learning~\cite{zhang2017amulet,zhang2018agile,zhang2020non} with multiple boundary-aware loss functions for each feature level.
The overall loss is defined as the summation over all side-outputs and the final fusion prediction:
\begin{equation}\label{12}
\mathcal{L} = \mathcal{L}^{f}+\sum^{S}_{s=1} \alpha_{s}\mathcal{L}^{s}
\end{equation}
where $\mathcal{L}^{f}$ is the loss of the final fusion prediction and $\mathcal{L}^{s}$ is the loss of the $s$-th side-output, respectively.
$S$ denotes the total number of the side-outputs and $\alpha_{s}$ is the weight for each level loss.
In this work, our method includes five side-outputs, \emph{i.e.}, $S=5$.

To obtain complete CD regions and regular CD boundaries, we define $\mathcal{L}^{f}$ or $\mathcal{L}^{s}$ as a combined loss with three terms:
\begin{equation}\label{13}
\mathcal{L}^{f/s} = \mathcal{L}_{WBCE}+\mathcal{L}_{SSIM}+\mathcal{L}_{SIoU},
\end{equation}
where $\mathcal{L}_{WBCE}$ is the weighted binary cross-entropy loss, $\mathcal{L}_{SSIM}$ is the structural similarity loss and $\mathcal{L}_{SIoU}$ is the soft intersection over union loss.
The $\mathcal{L}_{WBCE}$ provides a probabilistic measure of similarity between the prediction and ground truth from a pixel-level view.
The $\mathcal{L}_{SSIM}$ captures the structural information of change regions in patch-level.
The $\mathcal{L}_{SIoU}$ is inspired by measuring the similarity of two sets, and yields a global similarity in map-level.
More specifically, given the ground truth probability $g_{l}(\textbf{x})$ and the estimated probability $p_{l}(\textbf{x})$ at pixel $\textbf{x}$ to belong to the class $l$, the $\mathcal{L}_{WBCE}$ loss function is
\begin{equation}\label{14}
\mathcal{L}_{WBCE}=-\sum_{\textbf{x}}w(\textbf{x})g_{l}(\textbf{x})\text{log}(p_{l}(\textbf{x})).
\end{equation}
Here, we utilize weights $w(\textbf{x})$ to handle challenges appeared in CD: the class imbalance and the errors along CD boundaries.
Given the frequency $f_l$ of class $l$ in the training data, the indicator function $I$, the training prediction $P$, and the gradient operator $\nabla$, weights are defined as:
\begin{equation}\label{15}
w(\textbf{x})=\sum_{l}I(P(\textbf{x}==l))\frac{median(\textbf{f})}{f_l}+w_{0}I(|\nabla P(\textbf{x})|>0),
\end{equation}
where $\textbf{f} = [f_1,..., f_L]$ is the vector of all class frequencies.
The first term models median frequency balancing~\cite{badrinarayanan2017segnet} to handle the class imbalance problem by highlighting classes with low probability.
The second term assigns higher weights on the CD boundaries to emphasize on the correct prediction of boundaries.

The $\mathcal{L}_{SSIM}$ loss considers a local neighborhood of each pixel~\cite{wang2003multiscale}.
Let $\hat{\textbf{x}} = \{\textbf{x}_j: j = 1, ...,N^2\}$ and $\hat{\textbf{y}} = \{\textbf{y}_j: j = 1, ...,N^2\}$
be the pixel values of two corresponding patches (size: $N\times N$) cropped from the prediction $P$ and the ground truth $G$ respectively, the $\mathcal{L}_{SSIM}$ loss is defined as:
\begin{equation}\label{16}
\mathcal{L}_{SSIM}=1-\frac{(2\mu_{\textbf{x}}\mu_{\textbf{x}}+\epsilon)(2\sigma_{\textbf{xy}}+\epsilon)}{(\mu^{2}_{\textbf{x}}+\mu^{2}_{\textbf{y}}+\epsilon)(\sigma^{2}_{\textbf{x}}+\sigma^{2}_{\textbf{y}}+\epsilon)},
\end{equation}
where $\mu_{\textbf{x}}$, $\mu_{\textbf{y}}$ and $\sigma_{\textbf{x}}$, $\sigma_{\textbf{y}}$ are the mean and standard deviations
of $\hat{\textbf{x}}$ and $\hat{\textbf{y}}$ respectively.
$\sigma_{\textbf{xy}}$ is their covariance.
$\epsilon=10^{-4}$ is used to avoid dividing by zero.

In this work, one metric of interest at test time is the Intersection over Union (IoU).
Thus, we also introduce the soft IoU loss~\cite{mattyus2017deeproadmapper}, which is differentiable for learning.
The $\mathcal{L}_{SIoU}$ is defined as:
\begin{equation}\label{17}
\mathcal{L}_{SIoU}=1-\frac{\sum_{\textbf{x}}p_{l}(\textbf{x})g_{l}(\textbf{x})}{\sum_{\textbf{x}}[p_{l}(\textbf{x})+g_{l}(\textbf{x})-p_{l}(\textbf{x})g_{l}(\textbf{x})]}.
\end{equation}
When utilizing all above losses, the $\mathcal{L}_{WBCE}$ loss can relieve the imbalance problem for change pixels, the $\mathcal{L}_{SSIM}$ loss highlights the local structure of change boundaries, and the $\mathcal{L}_{SIoU}$ loss gives more focus on the change regions.
Thus, we can obtain better CD results and make the framework easier to optimize.
\section{Experiments}
\subsection{Datasets}
\textbf{LEVIR-CD}~\cite{chen2020spatial} is a public large-scale CD dataset.
It contains 637 remote sensing image pairs with a 1024$\times$1024 resolution (0.5m).
We follow its default dataset split, and crop original images into small patches of size 256$\times$256 with no overlapping.
Therefore, we obtain 7120/1024/2048 pairs of image patches for training/validation/test, respectively.

\textbf{WHU-CD}~\cite{ji2018fully} is a public building CD dataset.
It contains one pair of high-resolution (0.075m) aerial images of size 32507$\times$15354.
As no definite data split is widely-used, we crop the original image into small patches of size 256$\times$256 with
no overlap and randomly split it into three parts: 6096/762/762 for training/validation/test, respectively.

\textbf{SYSU-CD}~\cite{shi2021deeply} is also a public building CD dataset.
It contains 20000 pairs of high-resolution (0.5m) images of size 256$\times$256.
%
We follow its default dataset split for experiments.
There are 12000/4000/4000 pairs of image patches for training/validation/test, respectively.

\textbf{Google-CD}~\cite{liu2021super} is a very recent and public CD dataset.
It contains 19 image pairs, originating from Google Earth Map.
The image resolutions are ranging from 1006$\times$1168 pixels to 4936$\times$5224 pixels.
We crop the images into small patches of size 256$\times$256 with no overlap and randomly split it into three parts: 2504/313/313
for training/validation/test, respectively.
\subsection{Evaluation Metrics}
To verify the performance, we follow previous works~\cite{zhang2020deeply,bandara2022transformer} and mainly utilize F1 and Intersection over Union (IoU) scores with regard to the change-class as
the primary evaluation metrics.
Additionally, we also report the precision and recall of the change category and overall accuracy (OA).
\subsection{Implementation Details}
We perform experiments with the public MindSpore toolbox and one NVIDIA A30 GPU.
We used the mini-batch SGD algorithm to train our framework with an initial learning rate $10^{-3}$, moment 0.9 and weight decay 0.0005.
The batch size is set to 6.
For the Siamese feature extraction backbone, we adopt the Swin Transformer pre-trained on ImageNet-22k classification task~\cite{deng2009imagenet}.
To fit the input size of the pre-trained Swin Transformer, we uniformly resize image patches to 384$\times$384.
For other layers, we randomly initialize them and set the learning rate with 10 times than the initial learning rate.
We train the framework with 100 epochs.
The learning rate decreases to the 1/10 of the initial learning rate at every 20 epoch.
To improve the robustness, data augmentation is performed by random rotation and flipping of the input images.
For the loss function in the model training, the weight parameters of each level are set equally.
For model reproduction, the source code is released at https://github.com/AI-Zhpp/FTN.
\subsection{Comparisons with Sate-of-the-arts}
In this section, we compare the proposed method with other outstanding methods on four public CD datasets.
The experimental results fully verify the effectiveness of our proposed method.
\begin{table*}
\caption{Quantitative comparisons on LEVIR-CD and WHU-CD datasets.}
\label{tab:LEVIR}
\resizebox{1\textwidth}{!}
{
\centering
 \begin{tabular}{l|c|c|c|c|c|c|c|c|c|c|c|c|c|c|c|c|c|c}
 \hline
\multirow{3}{*}{Methods} & \multicolumn{5}{c|}{LEVIR-CD}&\multicolumn{5}{c|}{WHU-CD}\\
\cline{2-11}
~&Pre.&Rec.&F1&IoU&OA&Pre.&Rec.&F1&IoU&OA\\
\hline
\hline
FC-EF~\cite{daudt2018fully}        &86.91&80.17&83.40&71.53&98.39&71.63&67.25&69.37&53.11&97.61\\
FC-Siam-Diff~\cite{daudt2018fully}   &89.53&83.31&86.31&75.92&98.67&47.33&77.66&58.81&41.66&95.63\\
FC-Siam-Conc~\cite{daudt2018fully}   &91.99&76.77&83.69&71.96&98.49&60.88&73.58&66.63&49.95&97.04\\
BiDateNet~\cite{liu2020building}   &85.65&89.98&87.76&78.19&98.52&78.28&71.59&74.79&59.73&81.92\\
U-Net++MSOF~\cite{peng2019end}   &90.33&81.82&85.86&75.24&98.41&91.96&89.40&90.66&82.92&96.98\\
DTCDSCN~\cite{liu2020building}   &88.53&86.83&87.67&78.05&98.77&63.92&82.30&71.95&56.19&97.42\\
DASNet~\cite{liu2020building}   &80.76&79.53&79.91&74.65&94.32&68.14&73.03&70.50&54.41&97.29\\
STANet~\cite{chen2020spatial}   &83.81&\textbf{91.00}&87.26&77.40&98.66&79.37&85.50&82.32&69.95&98.52\\
MSTDSNet~\cite{song2022mstdsnet}   &85.52&90.84&88.10&78.73&98.56&-----&-----&-----&-----&-----\\
IFNet~\cite{zhang2020deeply}   &\textbf{94.02}&82.93&88.13&78.77&98.87&\textbf{96.91}&73.19&83.40&71.52&98.83\\
SNUNet~\cite{fang2021snunet}  &89.18&87.17&88.16&78.83&98.82&85.60&81.49&83.50&71.67&98.71\\
BIT~\cite{chen2021remote}     &89.24&89.37&89.31&80.68&98.92&86.64&81.48&83.98&72.39&98.75\\
H-TransCD~\cite{ke2022hybrid} &91.45&88.72&90.06&81.92&99.00&93.85&88.73&91.22&83.85&99.24\\
ChangeFormer~\cite{bandara2022transformer}   &92.05&88.80&90.40&82.48&99.04&91.83&88.02&89.88&81.63&99.12\\
Ours
&92.71&89.37&\textbf{91.01}&\textbf{83.51}&\textbf{99.06}&93.09&\textbf{91.24}&\textbf{92.16}&\textbf{85.45}&\textbf{99.37}\\
\hline
\end{tabular}
}
\end{table*}

\textbf{Quantitative Comparisons.}
We present the comparative results in Tab.~\ref{tab:LEVIR} and Tab.~\ref{tab:SYSU}.
The results show that our method delivers excellent performance.
More specifically, our method achieves the best F1 and IoU values of 91.01\% and 83.51\% on the LEVIR-CD dataset, respectively.
They are much better than previous methods.
Besides, compared with other Transformer-based methods, such as BIT~\cite{chen2021remote}, H-TransCD~\cite{ke2022hybrid} and ChangeFormer~\cite{bandara2022transformer}, our method shows consistent improvements in terms of all evaluation metrics.
On the WHU-CD dataset, our method shows significant improvement with the F1 and IoU values of 92.16\% and 85.45\%, respectively.
Compared with the second-best method, our method improves the F1 and IoU values by 0.9\% and 1.6\%, respectively.
On the SYSU-CD dataset, our method achieves the F1 and IoU values of 81.53\% and 68.82\%, respectively.
The SYSU-CD dataset includes more large-scale change regions.
We think the improvements are mainly based on the proposed DFE.
On the Google-CD dataset, our method shows much better results than compared methods.
In fact, our method achieves the F1 and IoU values of 85.58\% and 74.79\%, respectively.
We note that the Google-CD dataset is recently proposed and it is much challenging than other three datasets.
We also note that the performance of precision, recall and OA is not consistent in all methods.
Our method generally achieve better recall values than most methods.
The main reason may be that our method gives higher confidences to the change regions.
\begin{table*}
\caption{Quantitative comparisons on SYSU-CD and Google-CD datasets.}
\label{tab:SYSU}
\resizebox{1\textwidth}{!}
{
\centering
 \begin{tabular}{l|c|c|c|c|c|c|c|c|c|c|c|c|c|c|c|c|c|c}
 \hline
\multirow{3}{*}{Methods} & \multicolumn{5}{c|}{SYSU-CD}&\multicolumn{5}{c|}{Google-CD}\\
\cline{2-11}
~&Pre.&Rec.&F1&IoU&OA&Pre.&Rec.&F1&IoU&OA\\
\hline
\hline
FC-EF~\cite{daudt2018fully}        &74.32&75.84&75.07&60.09&86.02&80.81&64.39&71.67&55.85&85.85\\
FC-Siam-Diff~\cite{daudt2018fully}   &\textbf{89.13}&61.21&72.57&56.96&82.11&85.44&63.28&72.71&57.12&87.27\\
FC-Siam-Conc~\cite{daudt2018fully}   &82.54&71.03&76.35&61.75&86.17&82.07&64.73&72.38&56.71&84.56\\
BiDateNet~\cite{liu2020building}   &81.84&72.60&76.94&62.52&89.74&78.28&71.59&74.79&59.73&81.92\\
U-Net++MSOF~\cite{peng2019end}   &81.36&75.39&78.26&62.14&86.39&91.21&57.60&70.61&54.57&95.21\\
DASNet~\cite{liu2020building}   &68.14&70.01&69.14&60.65&80.14&71.01&44.85&54.98&37.91&90.87\\
STANet~\cite{chen2020spatial}   &70.76&\textbf{85.33}&77.37&63.09&87.96&\textbf{89.37}&65.02&75.27&60.35&82.58\\
DSAMNet~\cite{zhang2020deeply}   &74.81&81.86&78.18&64.18&89.22&72.12&80.37&76.02&61.32&94.93\\
MSTDSNet~\cite{song2022mstdsnet}   &79.91&80.76&80.33&67.13&90.67&-----&-----&-----&-----&-----\\
SRCDNet~\cite{liu2021super}   &75.54&81.06&78.20&64.21&89.34&83.74&71.49&77.13&62.77&83.18\\
BIT~\cite{chen2021remote}   &82.18&74.49&78.15&64.13&90.18&92.04&72.03&80.82&67.81&96.59\\
H-TransCD~\cite{ke2022hybrid}
&83.05&77.40&80.13&66.84&90.95&85.93&81.73&83.78&72.08&97.64\\
Ours
&86.86&76.82&\textbf{81.53}&\textbf{68.82}&\textbf{91.79}&86.99&\textbf{84.21}&\textbf{85.58}&\textbf{74.79}&\textbf{97.92}\\
\hline
\end{tabular}
}
\end{table*}

\begin{figure*}
\centering
\resizebox{1\textwidth}{!}
{
\begin{tabular}{@{}c@{}c@{}}
\includegraphics[width=1\linewidth,height=0.6\linewidth]{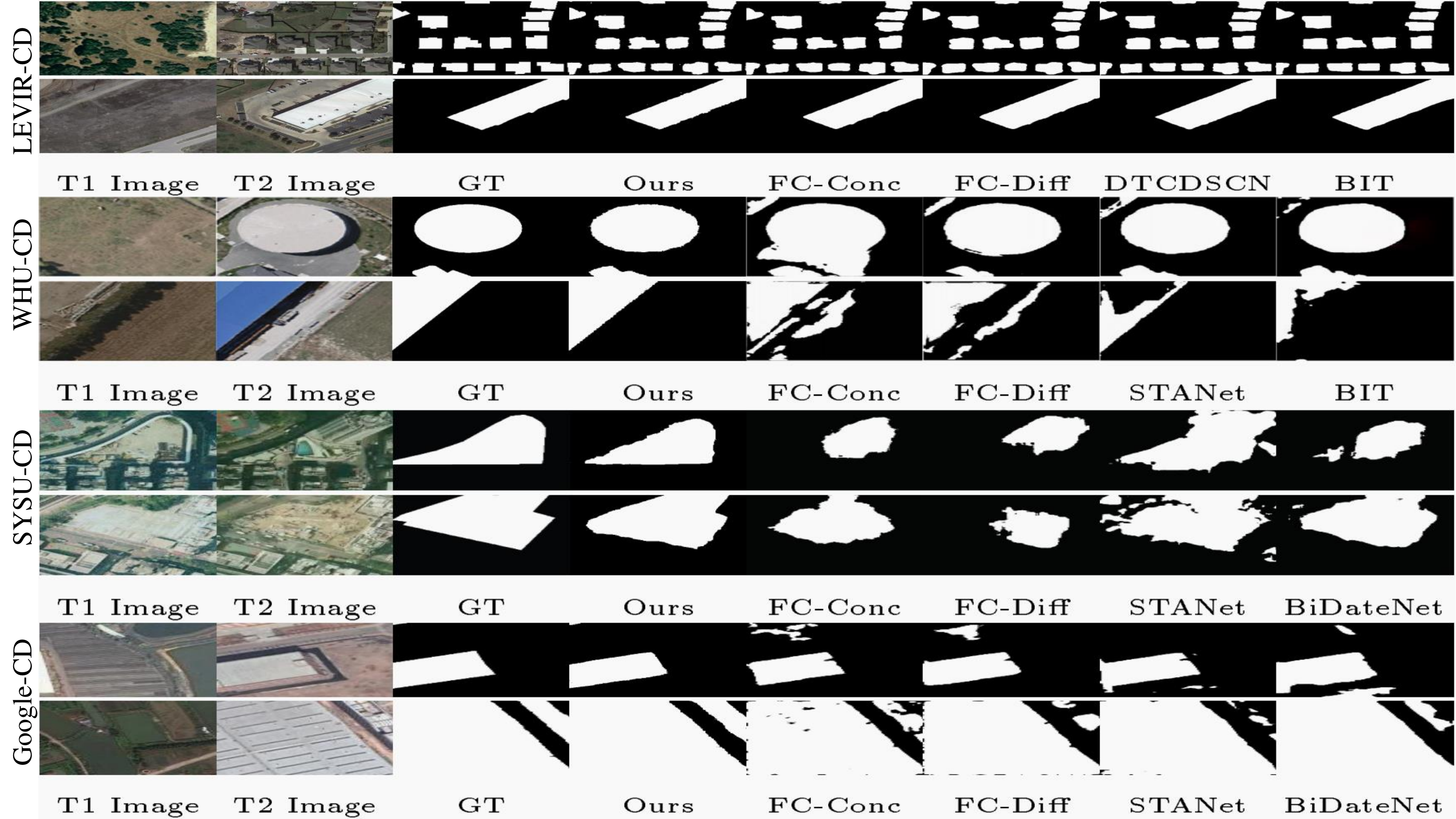} \\
\end{tabular}
}
\caption{Comparison of typical change detection results on four CD datasets.}
\label{fig:Comparsion}
\vspace{-6mm}
\end{figure*}

\textbf{Qualitative Comparisons.}
To illustrate the visual effect, we display some typical CD results on the four datasets, as shown in Fig.~\ref{fig:Comparsion}.
From the results, we can see that our method generally shows best CD results.
For example, when change regions have multiple scales, our method can correctly identify most of them, as shown in the first row.
When change objects cover most of the image regions, most of current methods can not detect them. However, our method can still detect them with clear boundaries, as shown in the second row.
In addition, when change regions appear in complex scenes, our method can maintain the contour shape.
While most of compared methods fail, as shown in the third row.
When distractors appear, our method can reduce the effect and correctly detect change regions, as shown in the fourth row.
From these visual results, we can see that our method shows superior performance than most methods.

\begin{figure*}
\vspace{-2mm}
\centering
\resizebox{1\textwidth}{!}
{
\begin{tabular}{@{}c@{}c@{}c@{}c@{}c@{}c@{}c@{}c@{}c@{}c@{}c@{}c}
\vspace{-0.8mm}
\includegraphics[width=0.1\linewidth,height=0.6cm]{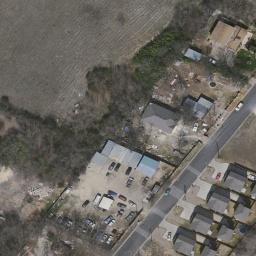}\ &
\includegraphics[width=0.1\linewidth,height=0.6cm]{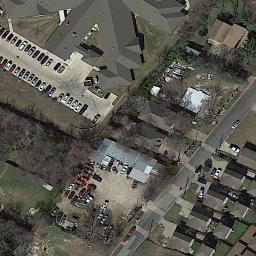}\ &
\includegraphics[width=0.1\linewidth,height=0.6cm]{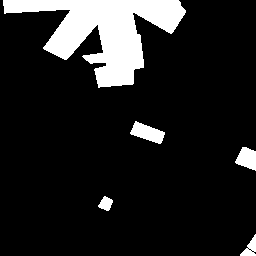}\ &
\includegraphics[width=0.1\linewidth,height=0.6cm]{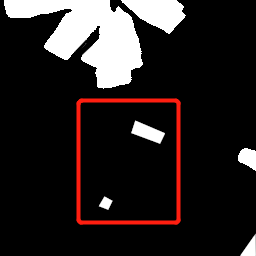}\ &
\includegraphics[width=0.1\linewidth,height=0.6cm]{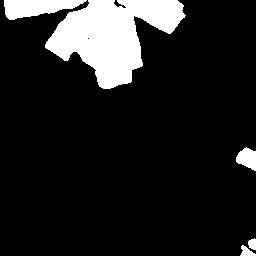}\ &
\includegraphics[width=0.1\linewidth,height=0.6cm]{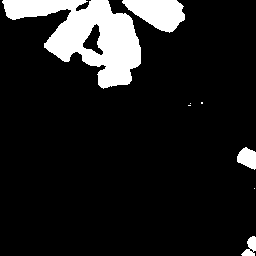}\ &
\includegraphics[width=0.1\linewidth,height=0.6cm]{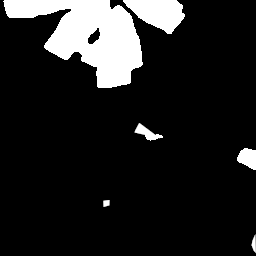}\ \\
\vspace{-0.8mm}
\includegraphics[width=0.1\linewidth,height=0.6cm]{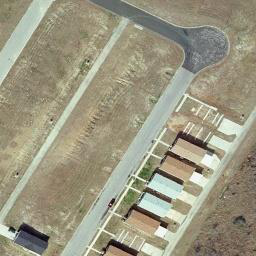}\ &
\includegraphics[width=0.1\linewidth,height=0.6cm]{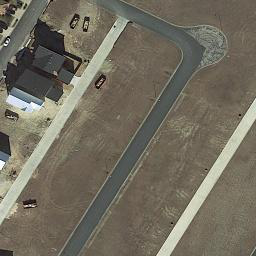}\ &
\includegraphics[width=0.1\linewidth,height=0.6cm]{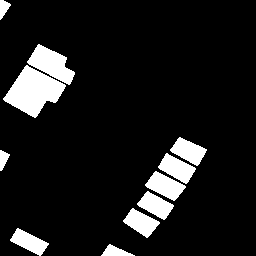}\ &
\includegraphics[width=0.1\linewidth,height=0.6cm]{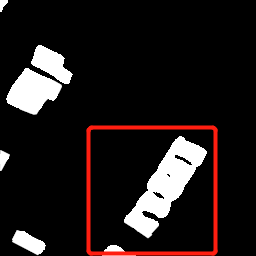}\ &
\includegraphics[width=0.1\linewidth,height=0.6cm]{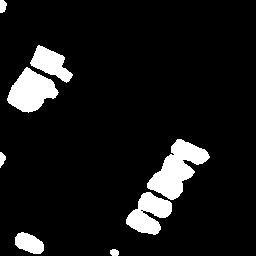}\ &
\includegraphics[width=0.1\linewidth,height=0.6cm]{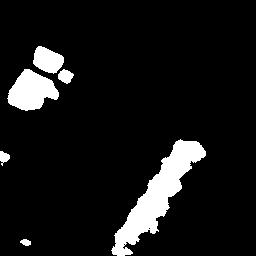}\ &
\includegraphics[width=0.1\linewidth,height=0.6cm]{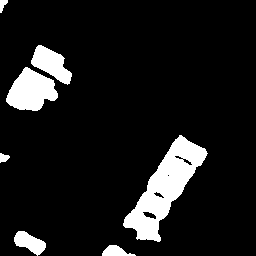}\ \\
\vspace{-0.8mm}
 {\tiny T1 Image} & {\tiny T2 Image} & {\tiny GT} & {\tiny Ours} & {\tiny DTCDSCN} & {\tiny IFNet} & {\tiny CFormer} \ \\
\end{tabular}
}
\caption{Comparison of typical change detection results on hard and failed samples.}
\label{fig:zoom}
\vspace{-8mm}
\end{figure*}

To further verify the effectiveness, we provide more hard samples in Fig.~\ref{fig:zoom}.
As can be seen, our method performs better than most methods (1st row).
Most of current methods can not detect the two small change regions in the center, while our method can accurately localize them.
Besides, we also show failed examples in the second row of Fig.~\ref{fig:zoom}.
As can be seen, all compared methods can not detect all the change regions.
However, our method shows more reasonable results.
\subsection{Ablation study}
In this subsection, we perform extensive ablation studies to verify the effect of key components in our framework.
The experiments are conducted on LEVIR-CD dataset. However, other datasets have similar performance trends.

\textbf{Effects of different Siamese backbones}.
As shown in the 2-3 rows of Tab.~\ref{tab:ablation1}, we introduce the VGGNet-16~\cite{simonyan2014very} and Swin Transformer as Siamese backbones.
To ensure a fair comparison, we utilize the basic Feature Pyramid (FP) structure~\cite{lin2017feature}.
From the results, one can see that the performance with the Swin Transformer can be consistently improved in terms of Recall, F1, IoU and OA.
The main reason is that the Swin Transformer has a better ability of modeling long-range dependency than VGGNet-16.
%
\begin{table}[htp]
\setlength{\tabcolsep}{5pt}
\centering
\caption{Performance comparisons with different model variants on LEVIR-CD.}
\resizebox{0.75\textwidth}{!}
{
\begin{tabular}{c|c|c|c|c|cc}
\hline
Models&Pre.&Rec.&F1&IoU&OA\\
\hline
(a) VGGNet-16+FP       &91.98&82.65&87.06&77.09&98.75\\
\hline
(b) SwinT+FP           &91.12&87.42&89.23&80.56&98.91\\
\hline
(c) SwinT+DFE+FP       &91.73&88.43&90.05&81.89&99.00\\
\hline
(d) SwinT+DFE+PCP      &92.71&89.37&91.01&83.51&99.06\\
\hline
\end{tabular}
}
\label{tab:ablation1}
\vspace{-4mm}
\end{table}

\textbf{Effects of DFE}.
The fourth row of Tab.~\ref{tab:ablation1} shows the effect of our proposed DFE.
When compared with the $Model (b)~SwinT+FP$, DFE improves the F1 value from 89.23\% to 90.05\%, and the IoU value from 80.56\% to 81.89\%, respectively.
The main reason is that our DFE considers the temporal information with feature summation and difference, which highlight change regions.

\begin{figure*}
\vspace{-2mm}
\centering
\resizebox{1\textwidth}{!}
{
\begin{tabular}{@{}c@{}c@{}c@{}c@{}c@{}c@{}c@{}c@{}c@{}c@{}c@{}c}
\vspace{-0.8mm}
\includegraphics[width=0.1\linewidth,height=0.62cm]{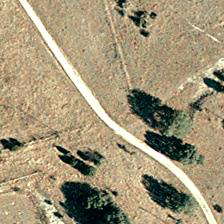}\ &
\includegraphics[width=0.1\linewidth,height=0.62cm]{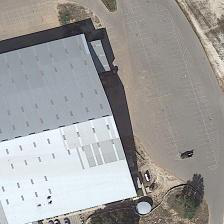}\ &
\includegraphics[width=0.1\linewidth,height=0.62cm]{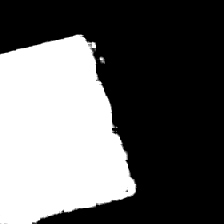}\ &
\includegraphics[width=0.1\linewidth,height=0.62cm]{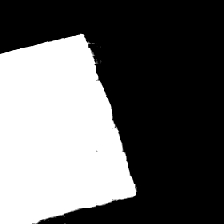}\ &
\includegraphics[width=0.1\linewidth,height=0.62cm]{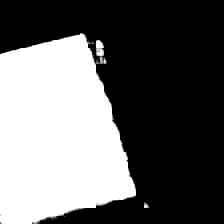}\ &
\includegraphics[width=0.1\linewidth,height=0.62cm]{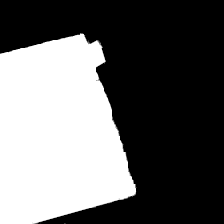}\ &
\includegraphics[width=0.1\linewidth,height=0.62cm]{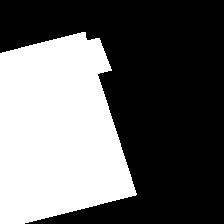}\ \\
\vspace{-0.8mm}
 {\tiny T1 Image} & {\tiny T2 Image} & {\tiny Model (a)} & {\tiny Model (b)} & {\tiny Model (c)} & {\tiny Model (d)} & {\tiny GT}\ \\
\end{tabular}
}
\caption{Visual comparisons of predicted change maps with different models.}
\label{fig:ablation}
\vspace{-4mm}
\end{figure*}

\textbf{Effects of PCP}.
In order to better detect multi-scale change regions, we introduce the PCP, which is a pyramid structure grafted with a PAM.
We compare it with FP.
From the results in the last row of Tab.~\ref{tab:ablation1}, one can see that our PCP achieves a significant improvement in all metrics.
Furthermore, adding the PCP also achieves a better visual effect, in which the extracted change regions are complete and the boundaries are regular, as shown in Fig.~\ref{fig:ablation}.

In addition, we also introduce the Swin Transformer blocks in the PCP as shown in Eq.~\ref{11}.
To verify the effect of different layers, we report the results in Tab.~\ref{tab:ablation2}.
From the results, we can see that the models show better results with equal layers.
The best results can be achieved with $n=4$.
With more layers, the computation is larger and the performance decreases in our framework.
\begin{table}[htp]
\setlength{\tabcolsep}{5pt}
\centering
\vspace{-2mm}
\caption{Performance comparisons with different decoder layers on LEVIR-CD.}\label{tab:ablation2}
\vspace{-2mm}
\resizebox{0.60\textwidth}{!}
{
\begin{tabular}{c|c|c|c|c|cc}
\hline
Layers&Pre.&Rec.&F1&IoU&OA\\
\hline
(2,2,2,2)      &91.18&87.00&89.04&80.24&98.90\\
\hline
(4,4,4,4)      &91.65&88.42&90.01&81.83&99.00\\
\hline
(6,6,6,6)      &91.70&88.30&89.96&81.76&98.99\\
\hline
(8,8,8,8)      &91.55&88.47&89.98&81.79&98.99\\
\hline
(2,4,6,8)      &92.13&85.71&88.80&79.86&98.89\\
\hline
\end{tabular}
}
\vspace{-4mm}
\end{table}

\textbf{Effects of different losses}.
In this work, we introduce multiple loss functions to improve the CD results.
Tab.~\ref{tab:ablation3} shows the effects of these losses.
It can be seen that using the WBCE loss can improve the F1 value from 88.75\% to 90.01\% and the IoU from 79.78\% to 81.83\%.
Using the SSIM loss achieves the F1 value of 90.11\% and the IoU of 82.27\%.
Using the SIoU loss achieves the F1 value of 91.01\% and the IoU of 83.51\%.
In fact, combining all of them can achieve the best results, which prove the effectiveness of all loss terms.
\begin{table}[htp]
\setlength{\tabcolsep}{5pt}
\centering
\vspace{-2mm}
\caption{Performance comparisons with different losses on LEVIR-CD.}
\label{tab:ablation3}
\vspace{-2mm}
\resizebox{0.7\textwidth}{!}
{
\begin{tabular}{c|c|c|c|c|cc}
\hline
Losses&Pre.&Rec.&F1&IoU&OA\\
\hline
BCE            &90.68&86.91&88.75&79.78&98.88\\
\hline
WBCE           &91.65&88.42&90.01&81.83&99.00\\
\hline
WBCE+SSIM      &91.71&88.57&90.11&82.27&99.01\\
\hline
WBCE+SSIM+SIoU &92.71&89.37&91.01&83.51&99.06\\
\hline
\end{tabular}
}
\vspace{-2mm}
\end{table}

\textbf{Scaling to higher resolutions.}
Since our work processes high-resolution remote sensing images, the scaling concern is very valuable.
In fact, most of compared methods utilize cropping for generating low-resolution input images.
We follow them and adopt a low resolution (256$\times$256), mainly considering the fairness of comparisons.
However, our work indeed can process a higher resolution with SwinT-Base/Small/Tiny.
Tab.~\ref{table:resolutions} shows the performance analysis with different resolutions and computation on WHU-CD.
One can see that our method can naturally scale to higher resolutions and show slightly better results.
\begin{table}
\vspace{-2mm}
\caption{Performance analysis with different input resolutions on WHU-CD.}
\label{table:resolutions}
\vspace{-2mm}
\begin{center}
\doublerulesep=0.1pt
\resizebox{0.70\textwidth}{!}
{
\begin{tabular}{|c|c|c|c|c|c|c|c|c|c|c|c|c|c|c|c|c|c|c|c|c|c|c|c|c|c|c|c|c|c|c|c|c|c|c|c|c|c|c|c|}
\hline
\multicolumn{4}{c|}{Input resolution}
&\multicolumn{4}{c}{Pre.}&\multicolumn{4}{c}{Rec.}
&\multicolumn{4}{c}{F1}&\multicolumn{4}{c}{IoU}
&\multicolumn{4}{c|}{OA}&\multicolumn{4}{c}{Flops(G)}
\\
\hline
\multicolumn{4}{c|}{256$\times$256}
&\multicolumn{4}{c}{93.09}&\multicolumn{4}{c}{91.24}
&\multicolumn{4}{c}{92.16}&\multicolumn{4}{c}{85.45}
&\multicolumn{4}{c|}{99.37}&\multicolumn{4}{c}{45}
\\
\multicolumn{4}{c|}{384$\times$384}
&\multicolumn{4}{c}{93.83}&\multicolumn{4}{c}{90.58}
&\multicolumn{4}{c}{92.17}&\multicolumn{4}{c}{85.48}
&\multicolumn{4}{c|}{99.37}&\multicolumn{4}{c}{134}
\\
\multicolumn{4}{c|}{512$\times$512}
&\multicolumn{4}{c}{94.21}&\multicolumn{4}{c}{90.25}
&\multicolumn{4}{c}{92.19}&\multicolumn{4}{c}{85.51}
&\multicolumn{4}{c|}{99.38}&\multicolumn{4}{c}{198}
\\
\hline
\end{tabular}
}
\vspace{-8mm}
\end{center}
\end{table}

\textbf{More structure discussions.}
There are some key differences between our work and previous fully Transformer structures: The works in~\cite{wu2021fully,he2022fully} are taking single images as inputs and using an encoder-decoder structure.
However, our framework utilizes a Siamese structure to process dual-phase images.
In order to fuse features from two encoder streams, we propose a pyramid structure grafted with a PAM for the final CD prediction.
Thus, apart from the input difference, our work progressively aggregates multi-level features for feature enhancement.
\section{Conclusion}
In this work, we propose a new learning framework named FTN for change detection of dual-phase remote sensing images.
Technically, we first utilizes a Siamese network with the pre-trained Swin Transformers to extract long-range dependency information.
Then, we introduce a pyramid structure to aggregate multi-level visual features, improving the feature representation ability.
Finally, we utilize the deeply-supervised learning with multiple loss functions for model training.
Extensive experiments on four public CD benchmarks demonstrate that our proposed framework shows better performances than most state-of-the-art methods.
In future works, we will explore more efficient structures of Transformers to reduce the computation and develop unsupervised or weakly-supervised methods to relieve the burden of remote sensing image labeling.
\bibliographystyle{splncs04}
\bibliography{egbib}
\end{document}